\documentclass[]{spie}  %>>> use for US letter paper
\usepackage{amsmath,amsfonts,amssymb}
\usepackage{graphicx}
\usepackage[colorlinks=true, allcolors=blue]{hyperref}
\usepackage{url}

\title{A real-time algorithm for human action recognition in RGB and thermal video}

\author{Hannes Fassold}
\author{Karlheinz Gutjahr}
\author{Anna Weber}
\author{Roland Perko}
\affil{JOANNEUM RESEARCH - DIGITAL\\
Steyrergasse 17, 8010 Graz, Austria \\
\emph{\{firstname\}.\{surname\}@joanneum.at}
}

\begin{document} 
\maketitle

\begin{abstract}
Monitoring the movement and actions of humans in video in real-time is an important task. We present a deep learning based algorithm for human action recognition for both RGB and thermal cameras. It is able to detect and track humans and recognize four basic actions (standing, walking, running, lying) in real-time on a notebook with a NVIDIA GPU. For this, it combines state of the art components for object detection (Scaled-YoloV4), optical flow (RAFT) and pose estimation (EvoSkeleton). Qualitative experiments on a set of tunnel videos show that the proposed algorithm works robustly for both RGB and thermal video.
\end{abstract}


\section{INTRODUCTION}
\label{sec:intro}  % \label{} allows reference to this section

% Introduction and maybe related work a bit, 1 page in total

Human action recognition deals with identifying the actions performed by a human from a video sequence, either in the visible domain (RGB) or a thermal video. It is an important task due to its many applications, like sports, healthcare,  security,  autonomous driving, ambient assisted living, human-machine interaction and video surveillance. In most of these application areas, the human actions must be detected in real-time in order to react accordingly. One exemplary scenario is subsurface structures, like subway stations or tunnels, which are indispensable for modern societies. To ensure safety and efficient reaction to a crisis, a deep understanding of the underground structure is necessary \cite{Perko2022, Perko2021} for specially trained and equipped personnel, aware of the associated risks and dangers.
 
In this work, we therefore propose an algorithm for real-time monitoring and action recognition of humans in subsurface structures. This is important especially in crisis situations (e.g. a mass panic in a subway station due to a power outage, explosion or terrorist attack), so that specially trained personnel is able to monitor the current situation and react accordingly. A power outage means of course also that all artificial lights in the subsurface structure are gone. So a video analysis relying on standard RGB cameras will not be able to monitor the situation and detect the actions of the people. Therefore, in addition to RGB cameras we rely also on thermal cameras, which are able to provide enough visual information (although at lower resolution than a RGB camera) in order to recognize the human actions.

 Our proposed algorithm detects four human actions: standing, walking, running and lying. The possibility to monitor these actions in real-time allows an operator to detect a potentially critical situation quickly and react in a proper way. For example, a person lying on the floor of a subway station usually indicates an emergency situation (like an injury). In the same way, a mass panic is usually characterized by many people running around and shouting. Also the information about where people are standing or walking in the station is very useful, e.g. for triggering an alarm when a person is crossing the line or is moving too close to the railway tracks.

For this, the algorithm relies on powerful deep learning based components for object detection and tracking (\emph{OmniTrack} and \emph{Scaled-YoloV4}), optical flow (\emph{RAFT}) and pose estimation (\emph{EvoSkeleton}). The paper is organized as follows. In section \ref{sec:components} we describe the key components which are employed within our algorithm. Section \ref{sec:actionreco} describes our proposed algorithm for real-time human action recognition, which relies on these components. Section \ref{sec:experiments} gives information about the qualitative experiments on a set of tunnel videos, which show that the proposed algorithm works robustly for both RGB and thermal video. Finally, section \ref{sec:conclusion} concludes the work and gives an outlook of our planned future work on this topic.
  
\section{Key components}
\label{sec:components}
The real-time algorithm for human action recognition (described in section \ref{sec:actionreco}) relies on a set of key components, specifically for the detection and tracking of persons in the video, for the estimation of their pose and on the calibration of the camera in order to gather a 3D representation of the scene. In the following, we will describe these key components more in detail.

\subsection{Object detection and tracking algorithm}
\label{subsec:detection_tracking}

For the detection and tracking of persons (and other objects), we base upon the \emph{OmniTrack} algorithm \cite{Fassold2019Omnitrack}. It is real-time capable and combines (in its original form) the YoloV3 \cite{Redmon2018YOLOv3AI} neural network for object detection with the TV-$L^1$ \cite{Wedel2008AnIA} optical flow algorithm for tracking.

In the following, the workflow of the object detection and tracking algorithm is described briefly. Let $I_{t-1}$ and $I_t$ be two consecutive frames for timepoints $t–1$ and $t$, and $S_{t-1}$ be the list of already existing scene objects which have been tracked so far until timepoint $t–1$. 
The workflow for each frame can be roughly divided into four phases: 
\begin{itemize}
%\begin{itemize}[leftmargin=0.2in]
  % from https://stackoverflow.com/questions/2611276/latex-beamer-way-to-change-the-bullet-indentation	
  %\setlength{\leftmargin}{2em}
\item In the \emph{preprocessing} phase, some general preparation steps (downscaling etc.) are done for the input image $I_t$ (the current frame of the video).
\item In the \emph{feature calculation} phase, the optical flow between the (downscaled) frames $I_{t-1}$ and $I_t$ is calculated, which gives a dense motion field $M_{t-1}$.  Additionally, the  object detector is invoked for $I_t$, yielding a list of detected objects $D_t$. 
\item In the \emph{prediction} phase, for all scene objects $S_{t-1}$ their motion is predicted in order to calculate their predicted position $S_t$ for timepoint $t$. 
\item In the \emph{matching} phase, a matching is done of the predicted scene objects $S_t$ against the detected objects $D_t$. The matching is done in a globally optimal way with the Hungarian algorithm~\cite{Kuhn1955}, and the intersection-over-union (IOU) metric is employed as the matching score.
\item In the \emph{update} phase, all scene objects in $S_t$ which could not be matched are considered as lost and a flag is set to exclude them from further processing. In contrast, all detected objects in $D_t$ which could not be matched are considered as newly appearing and therefore added to the scene objects list $S_t$. 
\end{itemize}

We updated the key components of the algorithm to more powerful methods which have been recently proposed. Specifically, for the object detector component we switched from YoloV3 to the \emph{Scaled-YoloV4} method \cite{Wang2020ScaledYOLOv4SC}. It achieves significantly higher accuracy by employing a cross-stage partial network and can be easily scaled to multiple resolutions. 

Additionally, instead of the classical TV-$L^1$ algorithm for optical flow we employ the recently proposed \emph{RAFT} optical flow algorithm \cite{Teed2020RAFTRA}. The RAFT optical flow method achieves high accuracy of the motion field and generalizes well to other domains (like thermal images which have a different characteristic than RGB images). 

Note that for both RGB and thermal input images, we use the standard Scaled-YoloV4 pretrained model, which has been trained on the MS COCO dataset \cite{lin2014microsoft} (consisting of RGB images). We do not fine-tune or retrain on a specific thermal image dataset. From the detection result we only use the detected objects of class \emph{human} and \emph{car}.

\subsection{Pose estimation algorithm}
\label{subsec:pose}

For human pose estimation, we employ the \emph{EvoSkeleton} algorithm \cite{Li2020CVPR}. The method evolves a limited dataset to synthesize unseen 3D human skeletons based on a hierarchical human representation and heuristics inspired by prior knowledge. Via this special data augmentation procedure, EvoSkeleton achieves state-of-the-art accuracy on the largest public benchmark (Human3.6M \cite{ionescu2013human}) and additionally generalizes well to unseen and rare poses. 

In order to calculate the poses (skeletons with 17 joints) for all detected persons in one frame, we proceed as follows. First, for all detected persons the rectangular regions of interest are extracted to a list of sub-images. These sub-images are now processed in multiple batches, with the size of the batch set to 4 sub-images. The batching mechanism makes inference more efficient and ensures that the GPU memory is not exhausted. With a batch size of 4, roughly 5 GB GPU RAM are occupied.

For thermal video, the color space of the input image is transformed, by converting it to greyscale and inverting it. This preprocessing of the image improves the result of the pose estimation (and also object detection), likely because the appearance of the transformed input image is nearer to the appearance of an image from a monochrome RGB camera. 

\begin{figure} [t]
   \begin{center}
   \begin{tabular}{c} 
   \includegraphics[height=8.5 cm]{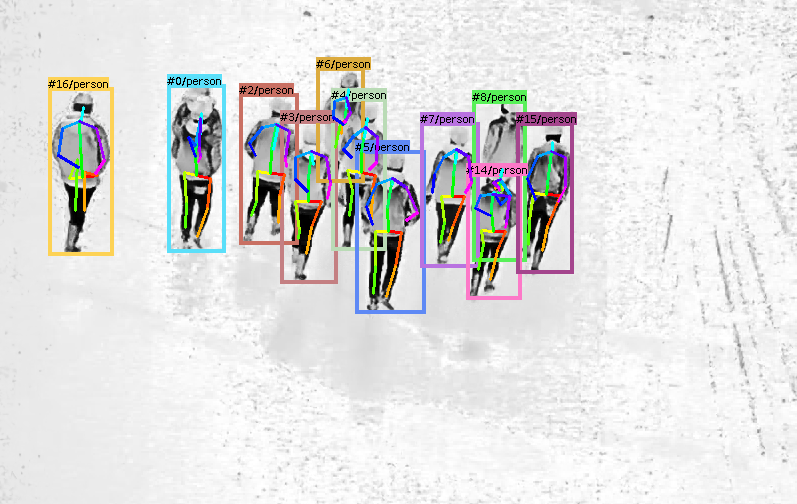}
	\end{tabular}
	\end{center}
   \caption[example] 
   { \label{fig:vis_thermal_skeleton_boundingboxes} 
Result of object detection (bounding boxes) and pose estimation (skeletons) for a thermal video.
}
\end{figure}

The pose estimation runs in parallel to object detection and tracking, so that real-time processing is ensured. For this, a separated process is spawned via the \emph{Multiprocessing} functionality in the Python programming language. The pose estimation runs at a reduced processing rate (roughly once per second), which is enough for the action recognition algorithm. See Figure \ref{fig:vis_thermal_skeleton_boundingboxes} for a result of the pose estimation (and object detection) for a thermal video.

\subsection{3D scene representation}
\label{scene_3d}

Since cameras only perceive a 2D projection of the 3D world, the location of a detected person cannot be directly projected into the 3D scene representation. Therefore, the cameras are calibrated intrinsically using a specially designed planar calibration random dot target. The materials are optimized such that the dark dots are visible for optical and thermals cameras (when heating the target by sunlight or an artificial light source). This calibration step determines the focal length, principal point, and lens distortion for each camera. The extrinsics are then determined using precisely measured ground control points within a least squares parameter adjustment, where the coarse camera location and orientation serve as starting point (cf. \cite{Perko2022}). 

The presented calibration allows 2D information in image geometry to be intersected with an existing 3D tube model that was acquired via terrestrial laser scanning for the whole subsurface test site (a street tunnel). Within this step for each pixel on the ground plane of the tunnel (roadway and sidewalk) the absolute 3D location is extracted in the subsurface reference projection. The latter information allows determining the velocity of tracked objects used in activity recognition. 

\section{Human action recognition algorithm}
\label{sec:actionreco}

The proposed real-time capable human action recognition algorithm relies on the components described in the previous section in order to identify and recognize characteristic actions of humans in the video, either from a RGB or thermal camera. Our proposed algorithm detects four human actions: \emph{standing, walking, running and lying}. The ability to monitor these actions in real-time allows an operator to detect a potentially critical situation quickly and react in a proper way. We briefly describe here how these human actions are identified.

We employ the OmniTrack algorithm described in section \ref{subsec:detection_tracking} in order to detect and track all humans which are visible in the video. The current speed of every human is also calculated over a certain timespan (e.g. the last three seconds), utilizing the 3D scene representation described in section \ref{scene_3d}. For all detected humans, their pose (skeletons with 17 joints) is estimated with the EvoSkeleton algorithm described in section \ref{subsec:pose}. From the edges of the skeleton corresponding to thorax and spine, a measure for the current orientation of the human's upper body is calculated. The measure is in degrees, in range $[0, 180]$. An orientation of approximately $90$ degrees indicates a vertically oriented upper body (i.e. a standing person), whereas an orientation of approximately zero or $180$ degrees indicates a horizontally oriented upper body (i.e. a lying person). 

Based on this metadata, we employ a rule-based system in order to identify the current action for each human. The rules for identification of a certain human action are straightforward:

\begin{itemize}
\item A human with a speed of more than $7$ km/h (kilometres per hour) is classified as \textbf{running}. This value is a good choice as it is stated in various resources as the average human running speed \footnote{\url{https://www.theglorun.com/what-is-the-average-running-speed-of-a-human/}}.
\item A human with a speed higher than $1$ km/h and lower than $7$ km/h is classified as \textbf{walking}.
\item A human with a speed of less than $1$ km/h and with a vertically oriented upper body is classified as \textbf{standing}. A vertically oriented upper body is detected if the calculated upper body orientation measure is between $65$ and $115$ degrees.
\item A human with a speed of less than $1$ km/h and with a horizontally oriented upper body is classified as \textbf{lying}. A horizontally oriented upper body is detected if the calculated upper body orientation measure is less than $65$, or more than $115$ degrees.
\end{itemize}

With this rule-based system, we identify an action for each human. As the measures (current speed and upper body orientation) are calculated over a certain timespan of a few seconds, the detection of the current human action has a small lag. For most applications, this does not pose an issue.

\begin{figure} [t]
   \begin{center}
   \begin{tabular}{c} 
   \includegraphics[height = 5.5 cm]{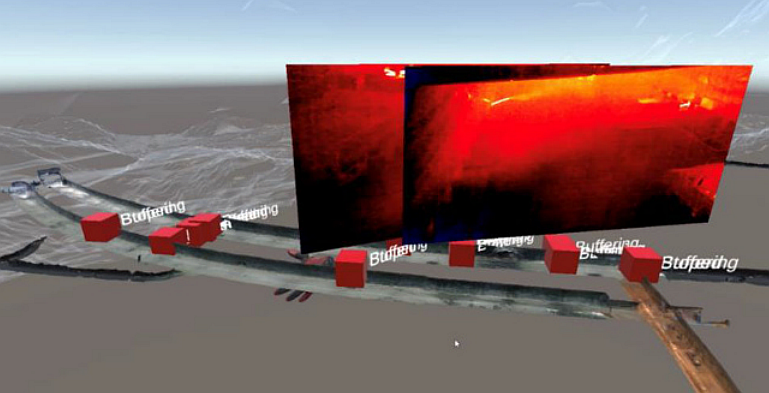}
	\end{tabular}
	\end{center}
   \caption[example] 
   { \label{fig:cop} 
Virtual Reality system for showing the algorithm result in real-time (image courtesy of \cite{Gegenhuber2022}).
}
\end{figure}

\section{Experiments and qualitative evaluation}
\label{sec:experiments}

As subsurface environment for testing of the proposed algorithm, the test site \emph{Zentrum am Berg} (ZaB) at the Styrian Erzberg in Austria was utilized \footnote{\url{https://www.zab.at/en/}}. It provides a unique and independent research infrastructure focussed on the operation of underground facilities. The underground facility consists of a twin-tube road tunnel and two parallel railway tunnel tubes as well as a test gallery. The facility is equipped with multiple optical and thermal cameras that serve as input for the developed computer vision system.

The experiments took place during the IRON NIKE research week \cite{Gegenhuber2022}, which was held from July 11, 2022 to July 14, 2022 at the ZaB and involved test persons from the Austrian Armed Forces and the Mine Rescue services. Several scenarios were simulated, like a terrorist attack in the tunnel and the rescue of injured people from a tunnel filled with smoke. The result of the proposed human action recognition algorithm in the tunnel was shown on a Virtual Reality system (see Figure \ref{fig:cop}) in order to get a \emph{common operational picture} of the situation in the current tunnels in real-time.

\begin{figure} [b]
   \begin{center}
   \begin{tabular}{c} 
   \includegraphics[height = 8.5 cm]{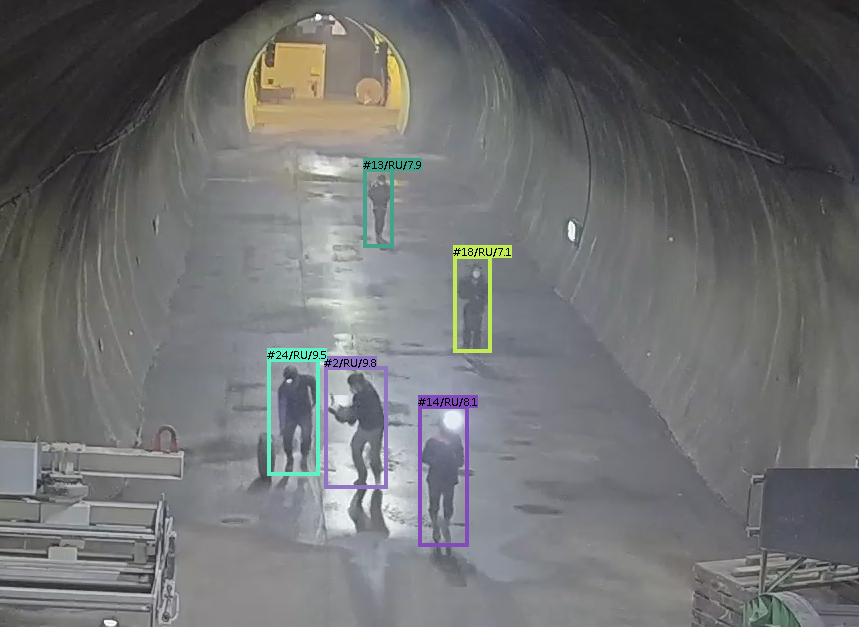}
	\end{tabular}
	\end{center}
   \caption[example] 
   { \label{fig:application_rgb} 
Visualisation of human action recognition for RGB video. First is the person ID, then the code for the human's action (WA - walking, RU - running, etc.), followed by the speed in km/h.
}
\end{figure}

\begin{figure} [t]
   \begin{center}
   \begin{tabular}{c} 
   \includegraphics[height = 8.8 cm]{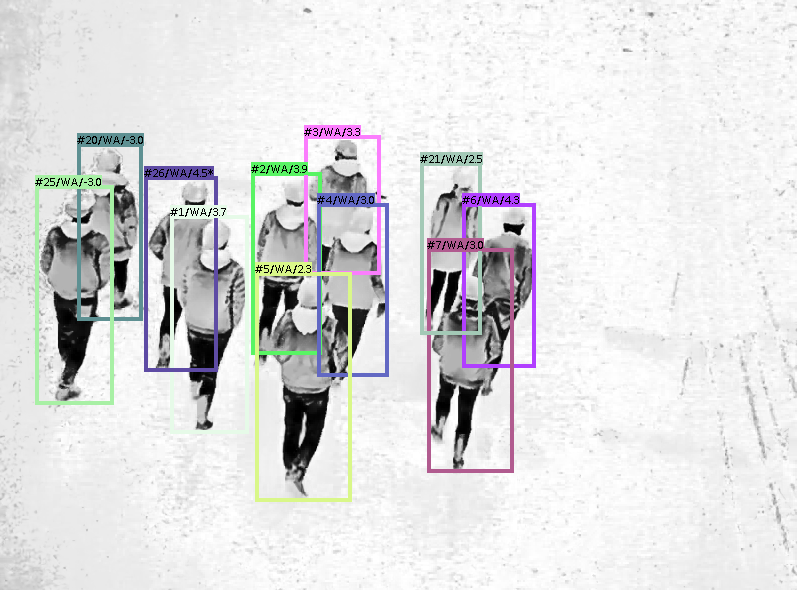}
	\end{tabular}
	\end{center}
   \caption[example] 
   { \label{fig:application_thermal} 
Visualisation of human action recognition for thermal video. First is the person ID, then the code for the human's action (WA - walking, RU - running, etc.), followed by the speed in km/h.
}
\end{figure}

For all simulated scenarios, the algorithm was able to detect and track the  humans reliably and to recognize their current action in real-time. See Figure \ref{fig:application_rgb} (RGB camera) and Figure \ref{fig:application_thermal} (thermal camera) for some visualizations. Even humans which are further away from the camera, and therefore are shown smaller in the video stream, could be detected reliably. The action recognition did also work quite well for the thermal cameras, there was no drop in the recognition performance noticeable compared to the results with RGB cameras.

The only action for which there is some room for improvement is the action "lying". This action is sometimes not recognized, especially if the human is further away from the camera. In this case, the employed object detector (Scaled-YoloV4) was not able to detect the lying person. If a human is not detected by the object detector, we cannot deduce any action for it.

The proposed algorithm runs in real-time on a Windows Notebook with a six-core CPU and a NVIDIA Geforce RTX 2070 with 8 GB GPU RAM. All deep learning based components, as well as the optical flow, run on the GPU. For each camera stream to be analyzed, a separate PC is needed.

\section{Conclusion}
\label{sec:conclusion}

A deep learning based algorithm for human action recognition was presented for both RGB and thermal cameras. It is able to detect and
track humans and recognize four basic actions (standing, walking, running, lying) in real-time. For this, it combines state of the art components for object detection, optical flow and pose estimation. Qualitative experiments in a realistic test situation show that the proposed algorithm works robustly for both RGB and thermal video.

In the future, we will work on expanding the set of human actions which are detected by the algorithm. For example, it could be useful to detect persons who are attacking other people physically or persons who are holding a gun. Furthermore, we will work on improving the recognition of the action "lying", e.g. by employing a more recent object detector like YoloV7 or by training a dedicated network which detects only humans.

\acknowledgments % equivalent to \section*{ACKNOWLEDGMENTS}       
 
The presented research activity is embedded into the projects NIKE-SubMovCon \#879720 (within the Austrian Security Research Programme KIRAS) and NIKE-DHQ Radiv \#886302 (within the Austrian Defense Research Programme FORTE), funded by the Austrian Research Promotion Agency (FFG).

% References
\bibliography{report} % bibliography data in report.bib
\bibliographystyle{spiebib} % makes bibtex use spiebib.bst

\end{document}